\documentclass[runningheads]{llncs}
\usepackage[T1]{fontenc}
\usepackage{graphicx}
\usepackage{tabularx}
\usepackage{amsmath}
\usepackage{float}
\begin{document}
\title{Machine vision-aware quality metrics for compressed image and video assessment}
%\titlerunning{Machine vision-aware quality metrics for compressed image and video assessment}
\author{Mikhail Dremin\inst{1}\orcidID{0009-0003-3572-6258} \and
Konstantin Kozhemyakov \inst{1}\orcidID{0009-0007-0605-2214} \and
Ivan Molodetskikh\inst{1}\orcidID{0000-0002-8294-0770} \and
Malakhov Kirill\inst{2} \and
Sagitov Artur\inst{2, 3} \and
Dmitriy Vatolin\inst{1}\orcidID{0000-0002-8893-9340}
}

\authorrunning{M. Dremin et al.}
\institute{Lomonosov Moscow State University \\
\email{\{mikhail.dremin, konstantin.kozhemiakov, ivan.molodetskikh, dmitriy\}@graphics.cs.msu.ru} \\
Huawei Technologies Co., Ltd. \\
Independent Researcher Linjianping
}
\maketitle              % typeset the header of the contribution

\begin{abstract}
A main goal in developing video-compression algorithms is to enhance human-perceived visual quality while maintaining file size. But modern video-analysis efforts such as detection and recognition, which are integral to video surveillance and autonomous vehicles, involve so much data that they necessitate machine-vision processing with minimal human intervention. In such cases, the video codec must be optimized for machine vision. This paper explores the effects of compression on detection and recognition algorithms (objects, faces, and license plates) and introduces novel full-reference image/video-quality metrics for each task, tailored to machine vision. Experimental results indicate our proposed metrics correlate better with the machine-vision results for the respective tasks than do existing image/video-quality metrics.
\end{abstract}
\vspace{-3mm}
\section{Introduction}

As the field of computer-vision continues to evolve, an increasing number of algorithms are being deployed in real-world applications. A popular application of this technology is video analytics, which has become integral to video surveillance, autonomous vehicles and other systems. Video analytics, for example, has two crucial tasks: detection and recognition. Depending on the system, the subjects of these tasks can be traffic signs, vehicles (object detection), human faces (detection/recognition), license plates (detection/recognition), and so on. As the number of video-surveillance cameras increases, automating these tasks becomes more critical given that human operators are incapable of processing such vast quantities of data.

To ensure efficient storage and transmission of such extensive data, the captured images and videos require compression. Lossy compression standards such as JPEG, H.264/AVC, H.265/HEVC, and AV1 serve this purpose; their development involved optimizing the visual quality of the compressed content. The best way to assess visual quality is subjective human ratings, but they can be time-consuming and expensive. Hence the use of full reference (FR) quality-assessment methods such as PSNR, SSIM~\cite{wang2004image}, and VMAF~\cite{vmaf}, some of which correlate highly with subjective scores~\cite{antsiferova}. These methods enable us to quickly and cost-effectively configure and develop codecs while emphasizing visual quality.

Most state-of-the-art detection and recognition algorithms are based on deep neural networks, and their effectiveness is evaluated not visually, but through performance metrics. Compression directly affects the performance and reliability of computer-vision algorithms, especially at high compression ratios~\cite{aqqa2019,byrne2022}. Vision researchers and developers have therefore attempted to determine image quality for algorithms and develop codecs for machine vision~\cite{mpegai}. 

For video surveillance system the main objects for analysis are people and vehicles, thus we choose three main video analytics algorithms for our machine-oriented quality metrics: object detection (including vehicles, persons, faces, and license plates), face recognition, license plates recognition.

Often video surveillance systems use a specific detection/recognition algorithm (e.g. YOLOv5). Optimizing camera-data compression for them requires a comparatively fast method that predicts detection/recognition performance on the already encoded video, thereby enabling selection of encoding parameters to maximize that performance. If the target neural-network-based detection/recognition algorithms are evaluating the relative performance of codec prototypes or codec settings, they need lots of time and computational power owing to the number of parameters and the neural-network size. For instance, the recent x264 codec has almost 50 settings; selection of these parameters through an exhaustive search, even without using complex neural networks, would take centuries~\cite{zvezdakov}.

During investigation for our machine oriented metric we have following targets:
\begin{enumerate}
\item Achieve high correlation score with mentioned three main video analytics algorithms for its particular implementations with lower computational complexity.
\item Considering question about metric generalization for different implementation detection/recognition algorithms.
\end{enumerate}

Additionally, detection/recognition algorithms are imperfect, and identifying the cause of potential errors is impossible. For example, an object could be truly unrecognizable in the encoded image or video, necessitating quality improvement, or one algorithm may have certain limitations whereas another can detect the object error-free. Running multiple detection/recognition algorithms to improve the robustness of such an evaluation method would be even more time-consuming and computationally intensive.

To address these issues, our paper makes three important contributions:
\begin{enumerate}
\item First, we propose a methodology of measuring image and video quality in terms of detection/recognition performance.
\item Second, we analyze detection- and recognition-performance correlation with that of popular image-quality-assessment (IQA) and video-quality-assessment (VQA) methods on widely used image and video codecs. Our results show little to no correlation between their outputs and detection/recognition performance.
\item Third, we propose new video-quality metrics based on convolutional-neural-network (CNN) models with respect to object detection, face detection/recognition, and license-plate detection/recognition performance. We validated our metrics by checking their correlation with the performance of the machine-vision algorithms for the corresponding tasks.
\end{enumerate}

\vspace{-3mm}
\section{Related work}
Methods for image-quality assessment have garnered considerable attention from researchers in the field of visual-perception as high-quality video content is important for retaining viewer interest~\cite{min2024}. Initially, these efforts evaluated image quality on the basis of how the human eye perceives visual information in generic videos and in specific video types such as streaming, user-generated content (UGC), 3D, and virtual and augmented reality. Recently, although computer vision increasingly permeates everyday life, efforts to develop machine-vision-aware image- and video-quality metrics have been less extensive. 
\subsection{Objective image- and video-quality assessment}
Many methods and algorithms evaluate the visual quality of images. Among the most widely used are PSNR and SSIM; they assess image quality on the basis of signal (pixel) similarity between the original and evaluated images. When applied to videos, these algorithms work frame by frame and then average the results. More-modern approaches to assessing video quality have emerged: for example, VMAF demonstrates a higher correlation with subjective human evaluations relative to PSNR and SSIM.

There are also no reference (NR) methods that take only a single image as input. Among them is an early NR quality metric: NIQE~\cite{mittal2012making}, which evaluates an image’s “naturalness” and serves in cases where the original image is unavailable. Recently, NR metrics have approached the quality of FR metrics. For instance, DOVER~\cite{wu2023exploring}and MDTVSFA~\cite{li2021unified} correlate highly with subjective quality~\cite{antsiferova}.

These methods are widely used to develop and optimize video-compression and video-processing algorithms. Some exhibit high correlation with human-perceived visual quality, but they were not designed to predict detection and recognition performance on compressed images and video.

\subsection{Image- and video-quality assessment for detection}
Kong et al.~\cite{kong2019blind} introduced a no-reference IQA algorithm for object detection. This algorithm integrates classical computer vision and selects 13 image features, including gradient-vector metrics and HOG descriptors~\cite{dalal2005histograms}. Random forest is trained to predict detection quality using the revised frame-detection-accuracy metric, which is akin to Intersection over Union (IoU).

Rahman et al.~\cite{rahman2021per} presented an algorithm that employs statistical features based on the internal representations of images input to a neural-network-based object detector. It then uses these statistics to train a LightGBM algorithm, which performs classification to predict whether the object-detection accuracy for a given input image will surpass a certain threshold or fail.

Schubert et al.~\cite{schubert2021metadetect} suggested predicting an object-detection algorithm’s accuracy on the basis of its confidence in the results. The authors analyzed non-maximum suppression step using features and statistics from the detection results to predict accuracy without ground-truth annotations. Their underlying hypothesis is that the more objects this stage filters out, the higher the confidence in the remaining object’s actual presence.

Beniwal et al.~\cite{beniwal2022image} proposed a metric based on the quantization error from H.264 compression. The mean DCT ratio of all filters of first Faster R-CNN convolutional layer is used as a quality label. Higher values indicate higher quantization loss and lower quality. They train CNN network to predict these quality labels using cropped patches as an input.

\subsection{Image- and video-quality assessment for recognition}
Best-Rowden and Jain~\cite{best2018learning} introduced an automatic method for predicting the quality of face images, integrating two assessment strategies: subjective evaluations by humans and objective measurement based on similarity scores for face recognition utility. Both approaches utilize features from deep neural networks as inputs for SVMs, allowing for a quantification of face image quality that reflects human perception and the operational performance of face-recognition systems.

Hernandez-Ortega et al.~\cite{hernandez2019faceqnet} introduced FaceQnet, a tool that estimates the quality of face images with respect to their utility in face recognition. FaceQnet operates by fine-tuning a preexisting face-recognition neural network for regression; the goal is to predict face-image quality as a continuous value.

Terhorst et al.~\cite{terhorst2020ser} proposed an unsupervised approach to face-image-quality estimation called SER-FIQ. They compute the quality score as the mean Euclidean distance of the multiple embedding features from a recognition model with different dropout patterns. 

Ou et al.~\cite{ou2021sdd} introduced SDD-FIQA, a method that uses inter- and intraclass comparison scores to determine the quality of face images by creating distributions of genuine and impostor scores for each image. The quality of an image is assessed by calculating the mean Wasserstein distance between these distributions across multiple iterations. This approach then refines a face-recognition model with these quality scores, similarly to the FaceQNet method.

Most of these studies consider quality loss due to shooting conditions—such as poor lighting, motion blur, and noise—but neglect artifacts that arise during compression with different quality factors. They also use knowledge of architecture and intermediate results of detection/recognition algorithms, although in some cases the algorithm is inaccessible.

\vspace{-3mm}
\section{Datasets}
Our research hinges on carefully collected datasets, each of which is pivotal to developing and testing image-quality metrics. We used the validation and/or test parts of popular datasets pertinent to the respective tasks: COCO 2017~\cite{lin2014microsoft} for object detection, WIDER FACE~\cite{yang2016wider} for face detection, CCPD~\cite{xu2018towards} for license-plate detection and recognition, and CelebA~\cite{liu2015deep} for face recognition. Our test set contained proprietary unlabeled videos from CCTV cameras for all tasks, except Glint360k images for face recognition.

\begin{figure}[h]
\includegraphics[width=\textwidth]{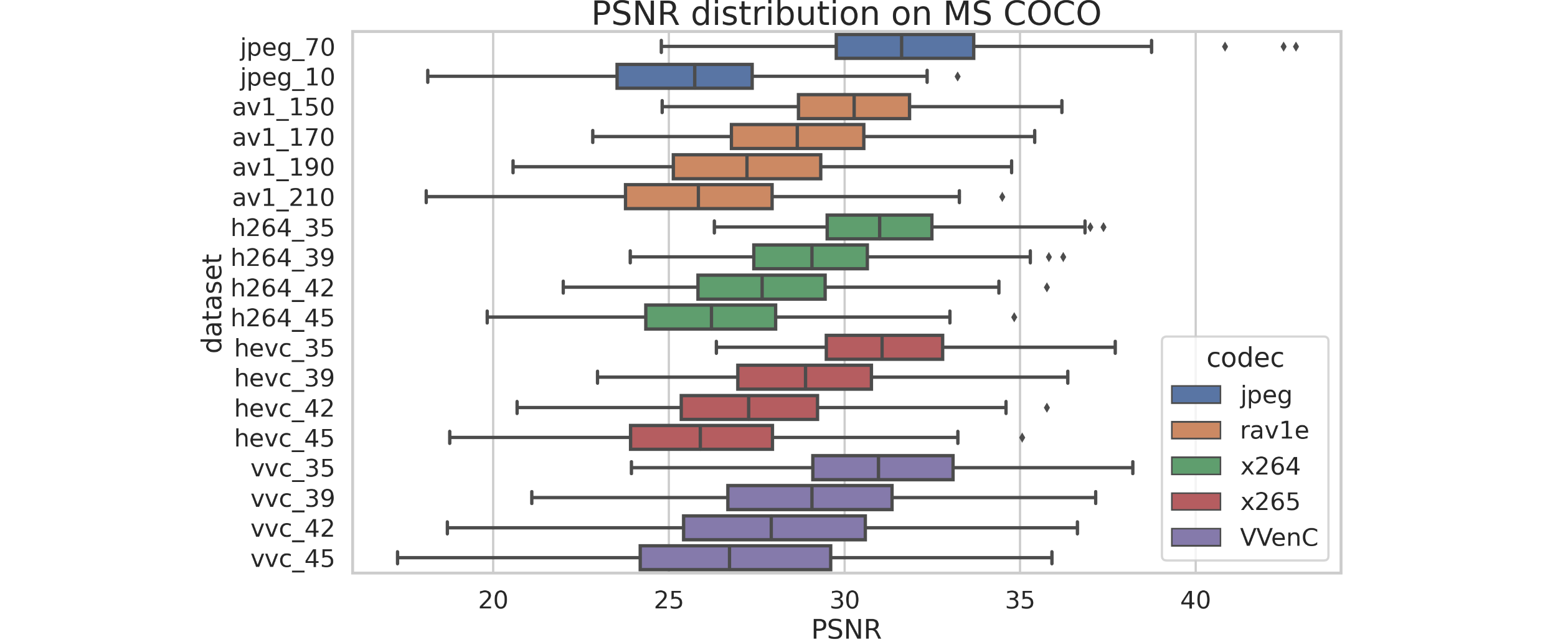}
\vspace*{-3mm}
\caption{PSNR distribution on MS COCO dataset, compressed with different codec and quality factors.}
\label{fig1}
\end{figure}

To ensure our method’s proper function despite various distortions, we selected several practical video and image codecs (rav1e, x264, x265, VVenC, and JPEG) to encode the dataset images. We balanced the dataset by calculating the PSNR of the compressed images and adjusting the codec quality parameters to achieve a similar distribution over all codecs, as Fig.~\ref{fig1} illustrates.

\vspace*{-3mm}

In total, every image in our datasets has 20 quality degradations. For JPEG, we used 20 compression degrees but retained only two quality factors because they represent two boundary values: the minimum, below which severe image degradation occurs, and the maximum, above which quality improvement is indiscernible. Note that we were unable to obtain completely uncompressed data; it would have been helpful, however, as open datasets usually employ JPEG compression and our proprietary videos employ H.264/H.265, so our compression distortions are on top of existing compression. This limitation should not significantly affect the results because open datasets have undergone quality control and avoid extreme compression, and because the camera codecs have a high bitrate.

\subsubsection{Test-dataset labeling}
The videos in our test set were unlabeled, but ground-truth labels are crucial for correct target-metric calculations, correlation-score measurement, and object extraction from images. We therefore used an automatic labeling pipeline in four of five tasks: object detection, face detection, license-plate detection, and license-plate recognition. 

\begin{figure}[h]
\includegraphics[width=\textwidth]{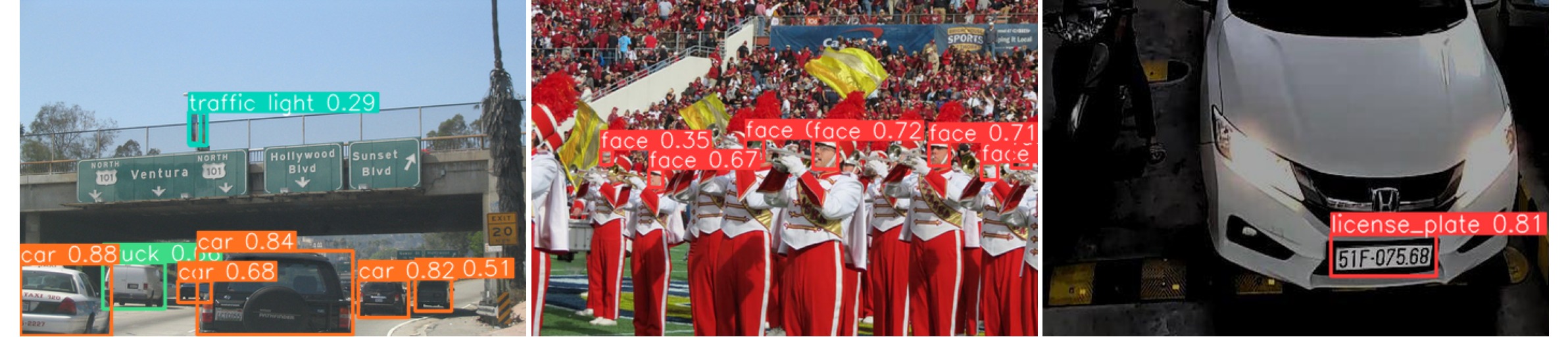}
\vspace*{-3mm}
\caption{Detection labeling example.}
\label{fig2}
\end{figure}

We extracted all the frames from videos and ran detection algorithms to label frames (each case used the algorithm’s most complex version for precise labeling: i.e., YOLOv5X for object detection, RetinaFace for face recognition, and LPRNet for license-plate recognition). We selected only objects with a corresponding high confidence and ignored detector errors. Instead, we looked at performance deterioration on compressed videos and picked about 1,000 frames containing several objects per task. Our choices were approximately 100 frames apart — since nearby frames are usually similar — to ensure diversity and reduce the number of noisy labels. We employed distinct frames (images), not videos, to train and validate results because video detection and recognition usually consider each frame and average the results for the entire video. Fig.~\ref{fig2}. show labeling examples. For license-plate detection we applied an extra filter using a recognition algorithm: it picked frames with fully recognized characters and filtered frames with similar license plates using the Levenshtein distance. The datasets were subsequently reviewed by humans for gross detection errors (false positive and false negative detections). The pixel-level accuracy of the bounding boxes was not meticulously verified: even though automatic annotation is often inaccurate, it is reasonable to expect that if a sophisticated detection algorithm for GT labeling fails to identify an object in an uncompressed image, the target detection algorithm will also struggle to detect the object once the image undergoes compression.

\vspace*{-3mm}

The face recognition test set was derived from the open Glint360k~\cite{an2021partial} dataset due to the need to extract at least two different face images for the same person. This requirement arose because our video recordings featured very similar facial images for each person, as individuals seldom appeared more than once in external video-surveillance footage. For each person in a dataset we choose two face images: one for the database and one for a query.

\begin{figure}[h]
\includegraphics[width=\textwidth]{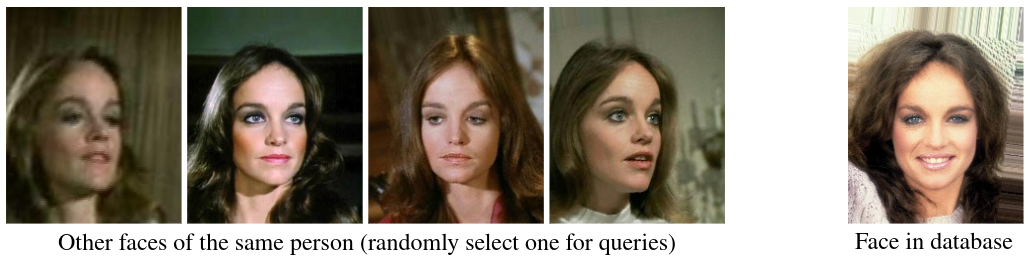}
\vspace*{-3mm}
\caption{Face images examples.}
\label{fig3}
\end{figure}

For the database we attempt to find the person’s “best” face image using ICAO-compliance software (Biolab-ICAO, as in FaceQNet’s pipeline). For a query we select another random image of that person’s face. Fig.~\ref{fig3} shows example images. Table \ref{tab1} summarizes the characteristics of our study’s datasets.

\vspace*{-3mm}

\begin{table}
\caption{Summary of datasets used in the study.}\label{tab1}
\begin{tabular}{|l|l|l|l|}
\hline
Task & Dataset part & Source images & Compressed images \\
\hline
Object detection & Train and val sets, COCO & 3,125 & 62,500 \\
\cline{2-4}
 & Test set, proprietary & 1,000 & 20,000 \\
\hline
Face detection & Train and val sets, WIDER & 3,226 & 64,520 \\
\cline{2-4}
 & Test set, proprietary & 1,000 & 20,000 \\
\hline
Car plate detection & Train and val sets, CCPD & 5,020 & 100,400 \\
\cline{2-4}
 & Test set, proprietary & 600 & 12,000 \\
\hline
Face recognition & Train and val sets, CelebA & 5,000 & 100,000 \\
\cline{2-4}
 & Test set, Glint360k & 1,000 & 20,000 \\
\hline
Car plate recognition & Train and val sets, CCPD & 5,020 & 100,400 \\
\cline{2-4}
 & Test set, proprietary & 600 & 12,000 \\
\hline
\end{tabular}
\end{table}

\vspace{-3mm}
\section{Proposed method}
\subsection{Detection methodology}
Object detection, crucial for applications like surveillance, identifies and locates objects in images by combining classification and localization. Key performance metrics include confidence score and Intersection over Union (IoU) with ground truth (GT). These metrics are essential for distinguishing between false negatives (missed detections) and false positives (incorrect detections), especially when objects are scarce. Correctly identifying missed detections is particularly important in surveillance, as human oversight can address false positives.

Object detectors tend to underperform in images that contain small or occluded objects~\cite{ren2015faster}, a limitation attributable to their implementation details rather than compression effects. Given our focus on information loss due to compression, it is essential to select quality metrics that are not biased by object-detector limitations.

The Mean Average Precision (mAP) metric assesses the detection algorithm’s confidence and incorporates an IoU threshold to determine the object-matching accuracy. However, its utility diminishes in images that contain few objects: here, mAP values may provide no meaningful insight because they may be binary (e.g., 0 or 1 for images with a single object).

Similarly, Average Precision (AP), which accounts for precision and recall, is poorly suited to single images or frames. The metric fails to differentiate between the effects of information loss (manifesting as false negatives) and detector inaccuracies (leading to false positives), particularly under varied compression levels.

We consider three possible target-detection performance metrics: mean-IoU, Object IoU, and Delta Object IoU. Mean-IoU provides a general measure of how accurately objects are detected over the entire image, regardless of object number or size. It permits consideration of false negatives: if no correct match to a reference object is found, the IoU for that object is zero. Furthermore, it allows setting of an IoU-value threshold below which objects are no longer considered correctly detected. Note, however, that although mean-IoU facilitates evaluation of individual images, it does not directly reflect the total number of objects, potentially obscuring detection performance relative to object quantity.

Object IoU focuses on the IoU for an object cropped from a reference frame. It provides a more specific measure of detection accuracy for individual objects by assessing how well the detected object matches the GT in size and location. Object IoU is particularly useful for analyzing detection performance object by object, a potentially critical capability for applications that must precisely detect each object.

Delta Object IoU represents the difference in an object’s IoUs between the reference frame and the compressed frame. Essentially, it quantifies the impact of compression on individual-object detection quality. A small Delta Object IoU indicates that the object’s detection accuracy is relatively unaffected by compression, whereas a large Delta Object IoU suggests considerable detection-performance degradation due to compression.

We then investigated which of our proposed target metrics is most representative for evaluating detection performance.
\subsection{Detection metric}
All target detectors in this study are YOLOv5 variations tailored to specific detection tasks: YOLOv5s for object detection, YOLO5Face for face detection, and LPD YOLOv5 for license-plate detection. First we analyzed almost 40 IQA and VQA metrics to determine how they correlate with the detectors’ performance. We applied each metric to every compressed image in our datasets.

Our analysis included the following:
\begin{itemize}
    \item Image/video-quality metrics
    \begin{itemize}
        \item Full-reference: PSNR (peak signal-to-noise ratio), SSIM (structural similarity index), MS-SSIM (multiscale structural similarity index), VMAF (video multi-method assessment fusion)
        \item No-reference: NIQE (natural image quality evaluator), BRISQUE (Blind/Referenceless image spatial quality evaluator)
    \end{itemize}
    \item Other Metrics: SAM (spectral angle mapper), SRE (spatial resolution enhancement), DSS (decision support system), NLPD (normalized Laplacian pyramid distance), GMSD (gradient magnitude similarity deviation), MDSI (mean deviation similarity index), VSI (visual saliency-induced Index), ERQA (edge-based region quality assessment)
    \item NSS (natural scene statistics): blockiness, total variation, colorfulness, brightness
\end{itemize}

\vspace*{-3mm}
\begin{figure}[h]
\includegraphics[width=\textwidth]{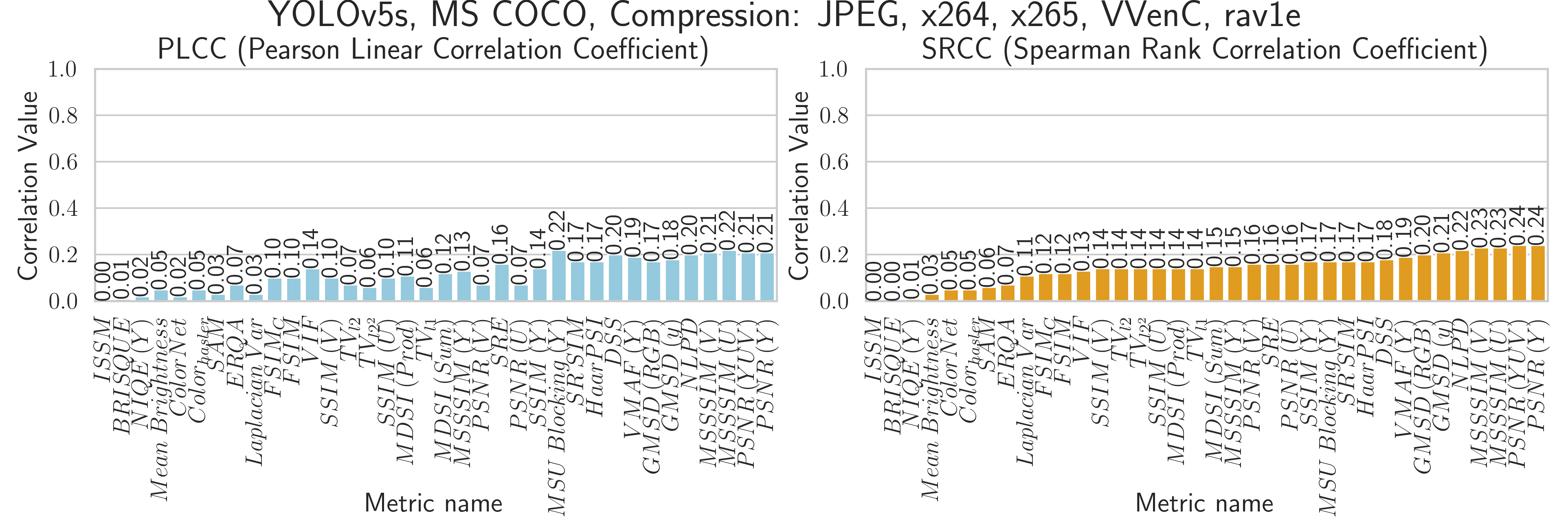}
\vspace*{-3mm}
\caption{Object detection. Objective metric results}
\label{fig4}
\end{figure}

After verifying the comparison method, we found the correlation scores for all tested metrics to be low, approximately 0.2–0.3 according to SRCC (see Fig.~\ref{fig4}, Table~\ref{tab2}). This result suggests none of the metrics are practical for evaluating detection performance on compressed videos.

\begin{table}[h]
\caption{Obtained correlations for standart IQA/VQA quality metrics.}\label{tab2}
\begin{tabularx}{\textwidth}{|X|X|p{2cm}|p{2cm}|p{2cm}|}
\hline
\multicolumn{3}{|c|}{Method} & \multicolumn{2}{p{4cm}|}{Best IQA/VQA metric \newline results among tested} \\
\hline
Task & Dataset & Target \newline algorithm & PLCC & SRCC \\
\hline
Object Detection & COCO & YOLOv5s & 0.21 & 0.24 \\
\hline
Face Detection & WIDER & YOLO5Face & 0.25 & 0.33 \\
\hline
Car plate Detection & CCPD & YOLO5Face & 0.25 & 0.33 \\
\hline
\end{tabularx}
\end{table}

\begin{figure}[h]
\includegraphics[width=\textwidth]{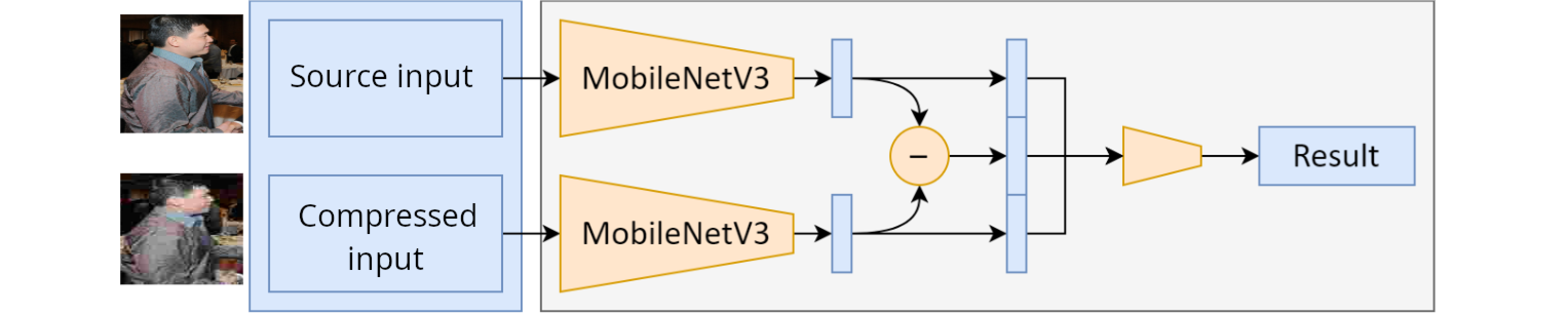}
\vspace*{-3mm}
\caption{The proposed quality metric architecture.}
\label{fig5}
\end{figure}

\begin{figure}[h]
\includegraphics[width=\textwidth]{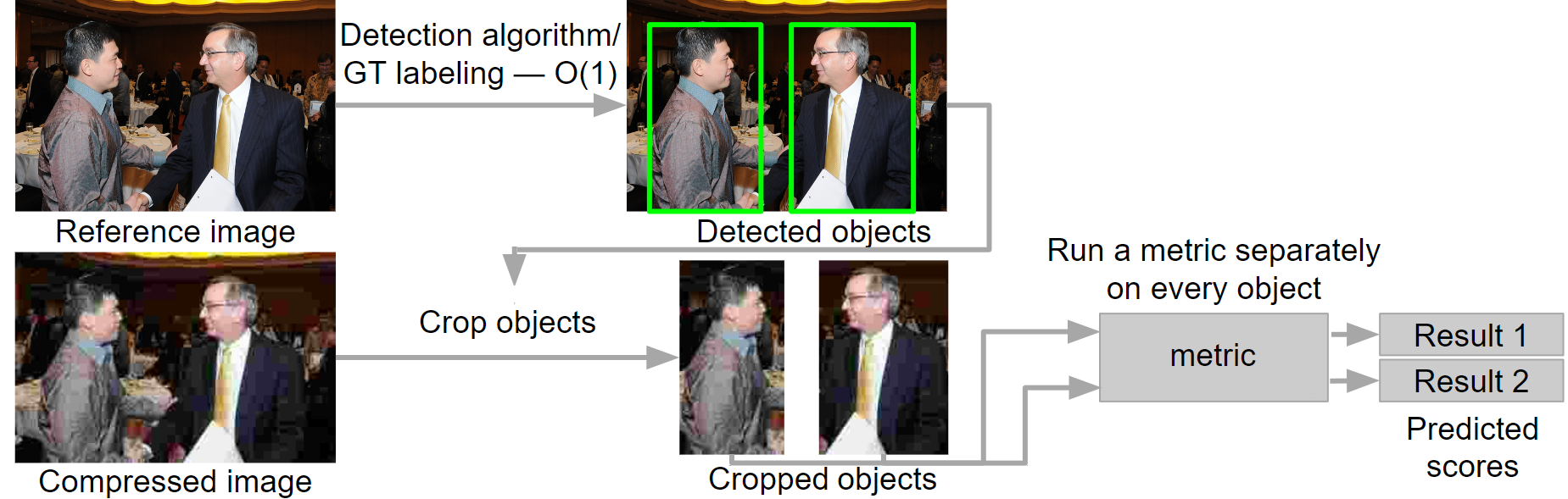}
\vspace*{-3mm}
\caption{The metric pipeline for assessing object crops.}
\label{fig6}
\end{figure}

Fig.~\ref{fig5} illustrates our proposed metric’s overall architecture. The input varies with the target metric: for mean-IoU, we used whole images; for object-IoU and Delta Object IoU, we used cropped objects based on GT bounding boxes (Fig.~\ref{fig6} shows this variant).

\vspace*{-3mm}

\vspace*{-3mm}

Due to IQA and VQA metrics' low correlation, we decided to research novel video-quality metrics based on CNN models which process crops of input images.

The lightweight MobileNetV3 network served as backbone for feature extraction. We then concatenated embeddings with their difference and fed the results to an MLP block that predicts the target metric.

We sequentially examined each proposed target-detection-performance metric to determine which was most representative. The first was mean-IoU. We trained our metric to predict the mean-IoU score for the entire compressed image as an input and achieved low SRCC scores of 0.29–0.37, depending on the task. Consequently, our analysis identified a major drawback of mean-IoU as the target performance metric. The IoU for each detected object in an image can vary, and averaging the results can lead to a false detection-quality perception in compressed content: the average value heavily masks omission of objects (false negatives). This drawback is absent from the next two metrics, which average the IoU of all objects in the dataset.

Having already calculated bounding boxes in uncompressed sequences (GT), we examined object IoU and computed the average value for cropped objects across all images. This approach recognizes that vast areas of a video frame, such as trees, asphalt, and sidewalks, are typically irrelevant to surveillance. By concentrating on areas that contain objects of interest, we sought to assess how the quality of these areas degrades rather than evaluating overall image or background quality. After training our metric to predict Object IoU as the target, we achieved an SRCC of 0.5–0.6 for detection. The results showed that the metric poorly predicted absolute IoU values. Indeed, predicting an IoU score for an object without any supporting information from the detector is difficult. Rather than considering compression’s impact on detection-algorithm performance, Object IoU considers the trained metric’s performance to predict the detection result, leading to misalignment with the actual objective.

To address described limitations, we shifted to the Delta Object IoU. That shift acknowledges the inherent detection-performance variability among compression levels and seeks to more directly quantify compression’s impact. Our experimental results demonstrate a notable improvement in correlation scores when using Delta Object IoU. Specifically, we observed SRCC in the 0.8–0.9 range for multiple tasks, indicating a stronger correlation with detection-algorithm performance degradation due to compression. Delta Object IoU’s main disadvantage as a target metric is that the deltas of different detection algorithms have low correlation, so the trained algorithm generalizes poorly to different target detectors.

Because of the unified approach of our proposed detection metrics, we can aggregate them into a general version. This version, trained on all datasets for all tasks, yielded results that were slightly inferior to those we obtained for task-specific metrics (see Table~\ref{tab3}).

\begin{table}[h]
\caption{Correlation scores (SRCC) for the general metric compared to individual metrics.}\label{tab3}
\centering
\begin{tabularx}{\textwidth}{|X|X|X|}
\hline
\textbf{Dataset (task)} & \multicolumn{2}{c|}{\textbf{Test data}} \\
\cline{2-3}
& \textbf{Generalized model} & \textbf{Separate models} \\
\hline
All datasets & 0.837 & --- \\
\hline
Object detection & 0.844 & 0.892 \\
\hline
Face detection & 0.816 & 0.818 \\
\hline
Car plate detection & 0.811 & 0.811 \\
\hline
\end{tabularx}
\end{table}

\subsection{Face-recognition methodology}
Standard metrics for measuring face-recognition efficiency include false-acceptance rate (FAR) and false-rejection rate (FRR). However, these can only be calculated for an entire dataset, whereas sometimes it is necessary to assess the quality of a single photo. A common methodology for neural-network-based face recognition involves computing embeddings for a corpus of reference images, each associated with known individuals, as well as for a query image. The identification process entails choosing the reference image whose embedding is most similar to that of the query image. To quantify the impact of image compression on face recognition performance, we calculated ArcFace~\cite{deng2019arcface} embeddings for all images (database and queries). Consider cosine similarity between image embeddings:

\begin{equation}
R(I_{a}, I_{b}) = cosSim(emb(I_{a}), emb(I_{b}))
\end{equation}

The proposed target metric is calculated as follows:

\begin{equation}
F(I_{ref}, I_{compr}) = R(I_{ref}, I_{database}) - R(I_{compr}, I_{database})
\end{equation}

Compression algorithms will influence correct recognition of the query face, so the cosine similarity between database and compressed images will differ. The face-recognition performance metric measures this difference in cosine similarity; if the difference is high, the face-recognition system is more likely to fail.

\subsection{Face-recognition metric}
\begin{figure}[h]
\includegraphics[width=\textwidth]{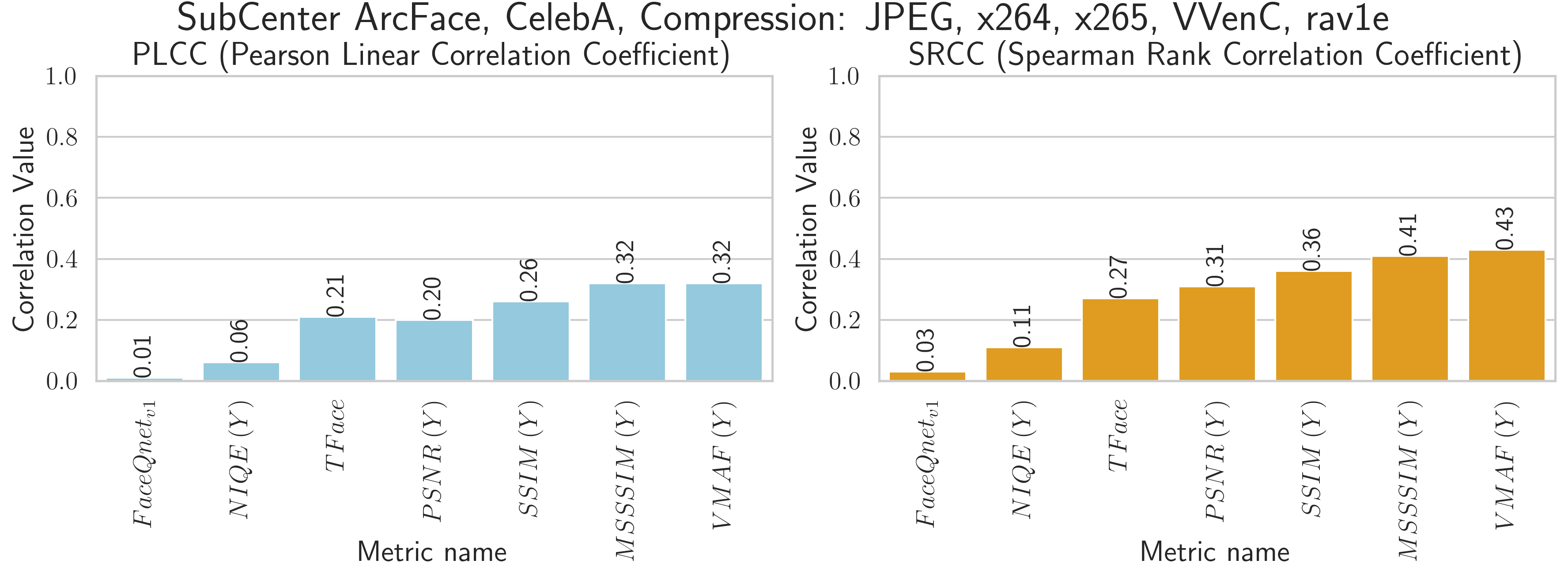}
\vspace*{-3mm}
\caption{Existing quality assessment methods for face recognition and standard IQA/VQA metrics results.}
\label{fig7}
\end{figure}

Recognition often follows detection, so we focused on image crops. As with detection, our first task was to analyze the performance of standard IQA/VQA metrics and existing quality-assessment methods for face recognition. The correlation scores for all metrics were low, suggesting that none is practical for evaluating face-recognition performance on compressed videos (Fig.~\ref{fig7}).

\vspace*{-3mm}

\vspace*{-3mm}
\begin{figure}[h]
\includegraphics[width=\textwidth]{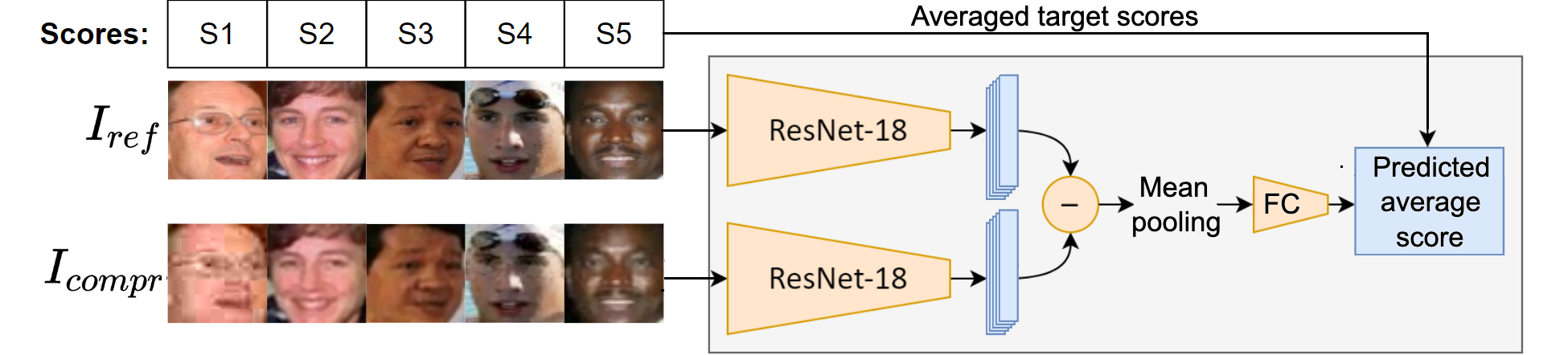}
\vspace*{-3mm}
\caption{Pipeline of face-recognition metric.}
\label{fig75}
\end{figure}

For face recognition we used ResNet-18 and a one-layer regression head to predict cosine-similarity deltas between source and compressed images. First, we trained our metric to predict a target score for a pair of reference and compressed face crops; the result was a 0.59 SRCC. We thus concluded that the metric provides unstable results on individual face images. To address this instability, we propose predicting image/video quality on small subsets of images as single inputs and averaging CNN feature before regression head to predict scores (Fig. ~\ref{fig75}). The SRCC of the metric trained with the proposed strategy is 0.85 SRCC.

\subsection{License-plate-recognition methodology}
To crop license plates from the full-size images, we used bounding boxes from GT. In license-plate recognition, the outcome is a sequence of characters, so the quality assessment differs fundamentally from other tasks considered. A variety of string-similarity measures, such as Hamming distance and Levenshtein distance, calculate the minimum number of character transformations necessary to convert one string to another. But these measures are unnormalized, making them less convenient for regression. Additionally, the Hamming distance requires equal-length strings, whereas license-plate-recognition methods may output a varying number of characters owing to unreadable symbols or misinterpretations, such as compression artifacts.

For this work, we propose Jaro similarity, which considers string lengths as well as the number and placement of common characters, making it suitable for automatic comparison and classification in data matching, duplicate identification, and more. The Jaro similarity $d_j$ of two given strings $s_1$ and $s_2$ is
\[ d_j = 
  \begin{cases} 
   0 & \text{if } m = 0 \\
   \frac{1}{3} \left( \frac{m}{|s_1|} + \frac{m}{|s_2|} + \frac{m-t}{m} \right) & \text{otherwise} 
  \end{cases}
\]
Where: $m$ is the number of \textit{matching characters}; $t$ is half the number of \textit{transpositions}.

It is particularly useful for license-plate recognition, as it accounts for matching characters that may be shifted a few positions left or right. Similar to IoU, the overall metric for an image comes from averaging Jaro similarity over all matched license plates.

\subsection{License-plate-recognition metric}
\begin{figure}[h]
\includegraphics[width=\textwidth]{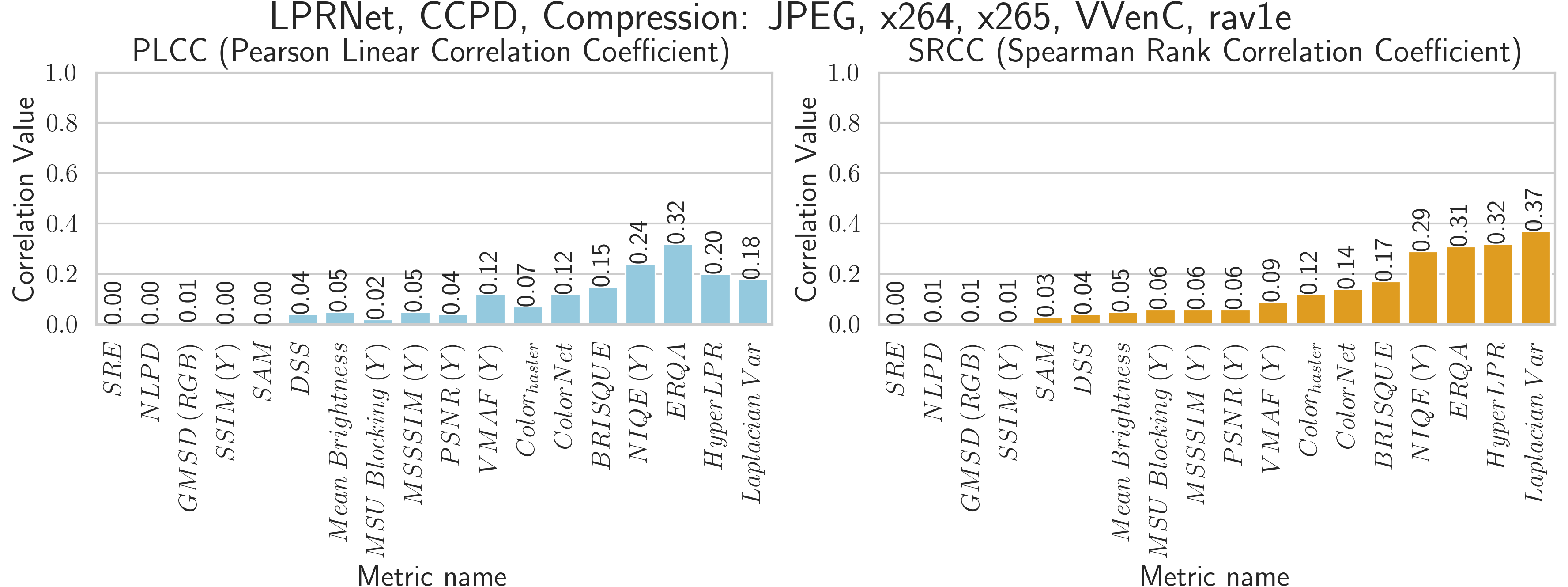}
\vspace*{-3mm}
\caption{Standard IQA/VQA metrics results.}
\label{fig8}
\end{figure}

The standard IQA/VQA methods exhibit bad correlation scores for our proposed target metric (Fig.~\ref{fig8}).

\vspace*{-3mm}

\vspace*{-3mm}
\begin{figure}[h]
\includegraphics[width=\textwidth]{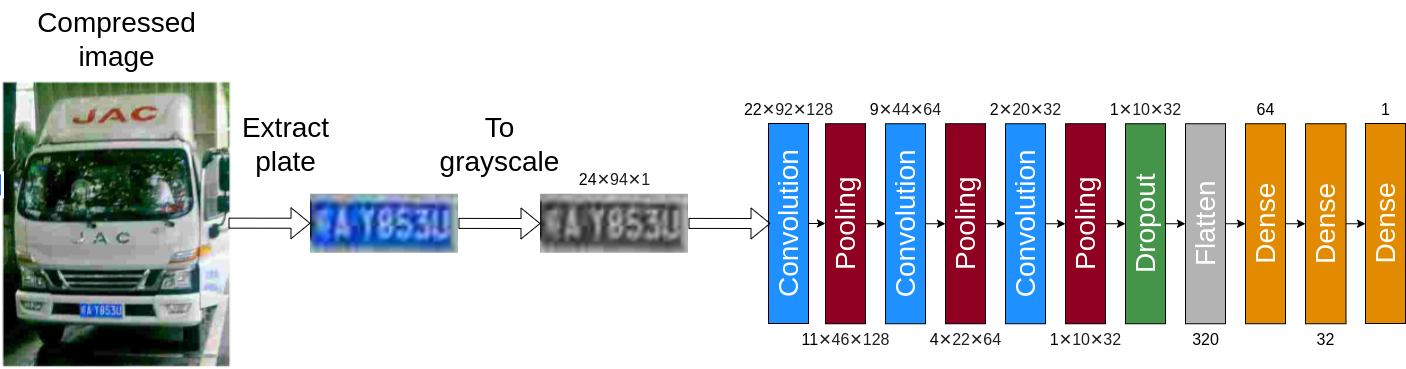}
\vspace*{-3mm}
\caption{Pipeline of license-plate-recognition metric.}
\label{fig9}
\end{figure}

We developed a neural-network metric to predict the performance of a license-plate-recognition method that takes as input a grayscale license-plate image. The model is a classic CNN, and its output is a single real number representing the expected recognition quality (Fig.~\ref{fig9}). A value of 1 indicates perfect recognition, while a value of 0 indicates the license-plate characters will be incorrectly recognized. After training the architecture, we achieved an SRCC of 0.85.

\vspace{-3mm}
\section{Conclusion}
This paper comprehensively explores the development and validation of novel image-quality metrics for machine vision, with a focus on assessing detection and recognition performance on compressed images and videos. These metrics address a major gap in standard IQA/VQA methods, which cater mainly to human visual perception and inadequately predict machine-vision performance on compressed images and videos.

Our metrics proved to be 3–5 times more computationally efficient than the target examined algorithms. For detection, we produced a general metric that applies to various detection subtasks, making the computations even more efficient since several detectors of the same family (e.g., YOLOv5) are used simultaneously in some cases (e. g., object and license-plate detection).

Although our proposed metrics yielded promising results, they are designed for specific tasks and, unmodified, may be inapplicable to other machine-vision tasks. Future work could explore development of more-general metrics that can handle a wider range of detection and recognition tasks. Additionally, further research could examine integration of these metrics into video-compression algorithms to automate optimization for machine vision.

Our metrics can also be used for other tasks that consume image/video data as an input. For example, visual question answering uses multimodal fusion of features extracted from both image and text. Taking a look at the existing VideoQA models, we can observe that the majority of them are utilizing features (rather than predicted labels) extracted from an image or a video by familiar object detection models, e.g. Faster-RCNN features in ViteVQA \cite{zhao2022towards} and BUTD~\cite{anderson2018bottom}, or ResNet features in MFH~\cite{Yu_2018}. Therefore, we expect that our results will extrapolate to those models. Applicability to newer models (e.g. Fuyu-8B) that use direct linear projection of image patches requires a separate study. As our machine-vision metrics are feature-focused for image/video data, they can serve as a reference point for developing a metric for the VisualQA task.

\vspace{-3mm}
\subsubsection{Acknowledgements}
This study was supported by Russian Science Foundation under grant 24-21-00172, https://rscf.ru/en/project/24-21-00172/

\vspace{-5mm}
\bibliographystyle{splncs04.bst}
\bibliography{references}{}
\end{document}